\documentclass[a4paper]{article}
\usepackage[utf8]{inputenc}
\usepackage[margin=1.08in]{geometry} 
\usepackage[parfill]{parskip}
\usepackage[hyphens]{url}
\usepackage{hyperref}
\usepackage[hyphenbreaks]{breakurl}
\usepackage[round]{natbib}
\usepackage{hyperref}
\usepackage{graphicx}
\hypersetup{colorlinks,linkcolor={blue},citecolor={blue},urlcolor={blue}}

\everypar{\looseness=-1}

\title{\vspace{-0.2cm} \bf Not Every AI Problem is a Data Problem:\\ We Should Be Intentional About Data Scaling}
\author{Tanya Rodchenko$^1$, Natasha Noy$^2$, Nino Scherrer$^3$ \vspace{1mm}\\ {\small $^1$Google, Technology \& Society, $^2$Google Research, $^3$Google, Paradigms of Intelligence}
\vspace{2mm}\\ \small{$^*$Opinions are our own and don’t represent the views of our employer}
}

\date{}

\begin{document}

\maketitle

\begin{abstract}
\noindent While Large Language Models require more and more data to train and scale, rather than looking for any data to acquire, we should consider what types of tasks are more likely to benefit from data scaling. We should be intentional in our data acquisition. We argue that the shape of the data itself, such as its compositional and structural patterns, informs which tasks to prioritize in data scaling, and shapes the development of the next generation of compute paradigms for tasks where data scaling is inefficient, or even insufficient.
\end{abstract}

\quad \vspace{4mm}\\
Large Language Models (LLMs) have revolutionized the AI landscape, demonstrating remarkable capabilities across a wide range of tasks. Each new model seemingly reinforces the notion that modern transformer-based AI can conquer any challenge if armed with sufficient compute and data. However, while scaling has accelerated certain applications, such as robotics, it has yet to show significant impact in others, such as identifying misinformation. We should not naively `throw' scaling at every problem. Instead, we should consider what types of problems are more likely to benefit from scaling, and in particular data scaling, and shift towards more intentional data acquisition.

We argue that the shape of data \citep{Carlsson2009TopologyAD} itself may hold valuable clues that could inform the success of data-driven scaling. For example, the presence of structural patterns and stability of data across multiple scales can help determine when data-driven scaling will be advantageous, and pinpoint where data scaling hits bounds. 

Moreover, the practicalities of data acquisition impose additional constraints that we must factor into the scaling equation upfront. Factors such as availability and verifiability of quality data, complexity and resource intensity of data collection, and availability of rigorous evaluation benchmarks determine not just the effectiveness but also the viability of data-driven scaling. 

\subsection*{Isn't Scaling All You Need?}
Scaling laws for LLMs have been driving our thinking on how to build optimal models almost since the inception of transformers \citep{vaswani2017attention}. In 2020, OpenAI researchers demonstrated the relationships between compute budget, model size, and dataset size for an optimal model \citep{kaplan2020scaling}. Subsequently, researchers from Google DeepMind \cite{hoffmann2022training} and other frontier labs investigated the compute-optimal growth of model and dataset size under a fixed compute budget, demonstrating that the model size and training data should be scaled equally. The scaling narrative suggests that this path could continue to yield improvements indefinitely, allowing LLMs to get better and better, and addressing more and more use cases over time.

However, most scaling laws rely on the inherent assumption that we have an infinite supply of quality data. In reality, quality data, and in particular human-generated quality data, is of course finite. As models grow into hundreds of billions of parameters, we tend to run out of quality data \citep{villalobos2022will, edwards2024data}.

Low quality data contains both irrelevant information like duplicates and diluted content, as well as unreliable or meaningless data. Some amount of low-quality data---assuming we can reliably identify it---is not only useful but also necessary: being able to distinguish between high-quality and low-quality data helps to improve the model's ability to detect and fix mistakes. However, training on large quantities of low-quality data—abundant on the web—without identifying it as such, is likely to lead to harmful effects on model performance and reliability \citep{lee2021deduplicating, longpre2023pretrainer, soldaini2024dolma}. Larger models are particularly sensitive to even small amounts of unreliable data. Their exceptional ability to memorize subtle patterns \citep{carlini2022quantifying} backfires: they memorize even one-off outliers, resulting in undesirable outputs such as the infamous `glue on pizza' response stemming from an ironic Reddit thread.\footnote{\url{https://www.theverge.com/2024/5/23/24162896/google-ai-overview-hallucinations-glue-in-pizza}} 

Synthetic data can help address some of the shortage of quality data in domains where we have the means to verify the quality of the generated data automatically \citep{liu2024bestpracticeslessonslearned}. Indeed, we can attribute some recent improvements in math and coding to synthetic data \citep{trinh2024solving, dubey2024llama}. However, math and coding are two fields where automatic verification is feasible (e.g.,\ by formalizing a math problem into a formal problem statement, and then tackling the problem until the verifier is satisfied). However, synthetic data is very unlikely to take the place of human-generated and sensor-generated data across most domains, as many real-world nuances are difficult to simulate \citep{liu2024bestpracticeslessonslearned}. Moreover, lack of diversity and quality control in the synthetic data generation pipeline may harm model robustness \citep{ liu2024bestpracticeslessonslearned} and introduce bias \citep{yu2024large}, and overdependence on it introduces risks of model collapse \citep{shumailov2024ai}. 

Together, these practical constraints of scaling laws challenge the notion that making models larger by simply adding more training data will lead to continuous and meaningful improvements across domains and tasks.

\subsection*{Where Data-Driven Scaling Thrives and Stumbles}
Let's consider the advancements that large models are catalyzing in robotics \citep{vanhoucke2024robotic}, drug discovery \citep{jumper2021highly}, and machine translation \citep{zhu2023multilingual}. What are the common threads among these success stories? 

Consider Machine Translation (MT). Early deterministic AI translation models struggled with contextual nuances of language \citep{ata2018position}, such as capturing culturally specific elements. Now transformer-based language models excel in this area due to several factors. First, the relatively static nature of language, with its abstractable rules and gradual vocabulary evolution, offered a stable foundation for model training. Second, high quality translation data, often sourced from reputable publishers and professional translations, further enabled effective training. Expansive datasets, including non-parallel corpora, enhanced the naturalness and contextual understanding of these models. Finally, LLM's vastly superior capacity to understand contexts (e.g., an entire document) led to significant improvements in translation quality. As a result, LLM-based techniques have brought such significant improvements to machine translation that they have become the new standard \cite{ciesielski2024translation}.

Robotics---from autonomous driving to factory automation---also demonstrates how data-driven scaling can outperform purely rule-based approaches. Vast sensor logs capture many of the scenarios and  environments where robots operate. The ``long tail'' of rare events, such as extreme conditions or erratic human behavior, demands richer inputs. Higher-resolution imagery and synthetic data increasingly address these demands. As sensor and camera costs drop and real-world data becomes more abundant, these conditions create a virtuous cycle of continuous improvement and propel robotics toward greater reliability than what was possible with prior methods.

Similarly, drug discovery benefits from a vast trove of past experiments, while fundamental biological processes remain relatively stable over time. However, coverage gaps, such as underexplored compound families, pose significant challenges. Pinpointing these gaps ensures that newly gathered data remains relevant and interpretable for model training. Meanwhile, stable features that emerge across experiments highlight critical research directions, enabling more targeted and intentional data acquisition.

At the other end of the spectrum is one of AI's most formidable challenges---robust and reliable reasoning, where AI systems often fail surprisingly at tasks that are easy for humans. At the time of writing, Google’s Gemini 2.0\footnote{\url{https://ai.google.dev/gemini-api/docs/thinking-mode}} and OpenAI’s o1\footnote{\url{https://openai.com/o1/}} have just raised the bar on reasoning benchmarks. And yet, OpenAI’s o1 still fails on simple alterations of existing benchmarks \citep{mirzadeh2024gsm, gulati2024putnam} or more complex math problems \citep{glazer2024frontiermath}.\footnote{Evaluating Google’s Gemini 2.0 on these particular benchmarks is still pending.} In our discussions with various field experts, they almost unanimously agreed that this challenge is primarily rooted in model architecture and learning algorithms, rather than data-driven scaling. 

\subsection*{Predictive Power of Data Shape}
Successful use cases offer clues on where data scaling will be helpful under the current learning paradigm. We find the framework of Topological Data Analysis \citep{Carlsson2009TopologyAD}, which aims to identify the intrinsic dimensions and patterns within datasets, particularly useful. While scientists discussed the concept of ‘shape of data’ as early as 2009, it still remains relatively underexplored \citep{uchendu2024unveiling}.  

Topological features derived from data, such as data’s compositional and structural patterns, and their evolution over time, provide insights into whether certain applications are suitable for data-driven scaling. In applications that require an understanding of connectivity and higher-order relationships within a dataset, data shape can reveal stable structures across multiple scales (i.e. across different levels of granularity or abstraction) \citep{wu2021topology}. For example, translation between languages exhibits regular and persistent patterns at different scales (across sentences, paragraphs, documents). In general, language patterns are stable over time. We know what type of data we need to expand to new languages. 
And while it may be challenging to acquire the data for rare or only spoken languages, it is easy to judge whether newly acquired data is what we need.

In contrast, use cases where data lacks strong, persistent topological features or where the structure is highly fragmented or unstable over time, may not be as well-suited for data scaling approaches. In particular, tasks that involve noisy, unstructured, or random data, where no clear topological or geometric patterns emerge, can be more challenging for models to handle effectively.

Journalistic fact checking and exposing misinformation is one such example. LLMs had success on some fact checking tasks, primarily because they understand language better and can match the text against sources in a more flexible way, particularly when grounded with real-time search data. However, new misinformation techniques evolve rapidly, rendering earlier patterns less helpful. The almost infinite number of types of misinformation makes the scope of the problem formidable, even for a human fact checker. It is not clear what data will effectively predict future misinformation. Uncertainty management---critical for navigating complex `grey area' fact checking tasks---also remains a largely unsolved problem from a technical standpoint, with limited benefit to gain from more data. Finally, understanding what information is reliable is both intuitive and contextual, and sometimes even culturally specific or subjective. Researchers are actively working on many approaches to identify misinformation, but they are less focused on collecting more data as a sole solution.

\subsection*{Data Acquisition as Another Predictor}
Beyond data shape, the feasibility of data-driven scaling is  largely determined by the nature of the data-acquisition process. Intuitively, if quality data is available and accessible, the potential for scaling increases significantly. For example, as we continue to collect sensor data from autonomous cars, their reliability and performance will continue to improve.

The specific type of data to collect also plays a crucial role. For example, training on step-by-step--style content that embodies ``procedural knowledge'' leads to significant performance improvement in the current learning paradigm~\citep{ruis2024procedural, yu2024distilling}. The abundance of such data in domains like math and coding, coupled with the emergence of strong evaluation metrics, has fueled rapid progress in these areas. However, acquiring or creating procedural knowledge for other tasks remains a significant challenge.

As we discuss acquiring quality data, it is crucial to note that the very definition of data quality is nuanced. For starters, quality data needs to be accurate, reliable, and complete. Additionally, training data must be dense with useful information and offer novel insights compared to the data already available for training. High quality data needs to be at the right level of granularity given the use case, timely and diverse enough to provide meaningful insights. In other words, the definition of high quality data is not universal; the quality is connected to a use case and the value that the trained model delivers to the user. 

Finally, we cannot fully assess the impact of data without critically looking at today's evaluation frameworks. Our evaluation approaches need to better reflect the nuances and needs of real-world users, and focus less on relatively simplistic single-turn benchmarks. Current benchmarks test limited aspects of performance which do not necessarily translate to user value. The next generation of evaluation approaches needs to consider how AI models handle real-world complexities, measure stochastic performance, and reflect user satisfaction and economic value.

Understanding the data-acquisition challenges is a critical component of making an informed decision about a potential scaling initiative, including an ultimate assessment of whether the benefits of scaling outweigh the costs.

\begin{figure}[t!]
    \centering
    \vspace{-1.2cm}
    \includegraphics[width=1.0\textwidth]{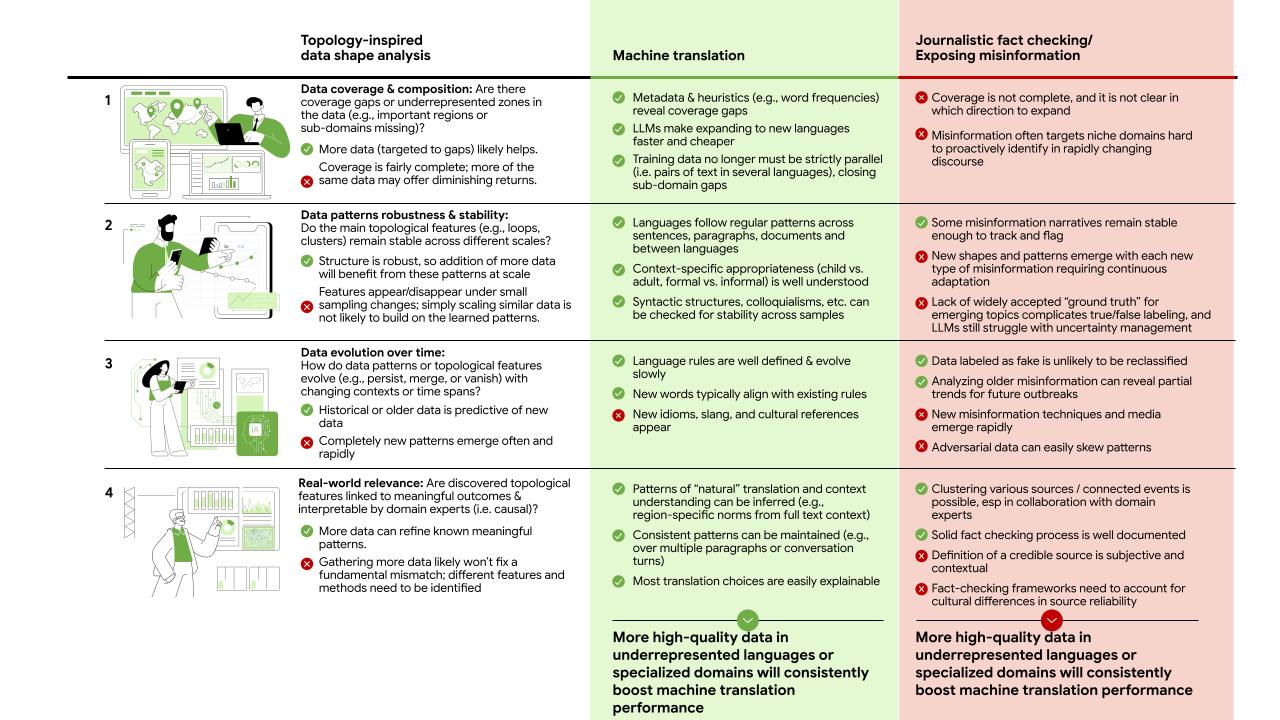}
    \caption{\textbf{Operationalizing data-shape based assessment of use cases}: The table lists the types of questions that will help understand the shape of data for specific use cases. We demonstrate the application of this framework to two use cases, machine translation and understanding misinformation.}
\end{figure}

\subsection*{The Promise of Intentional Data Scaling}

We should be intentional in data-driven scaling. By focusing on use cases with a strong hypothesis about efficacy of scaling, and by collecting fit-for-purpose data based on the needs of these use cases, we can make model training more efficient and reduce the volume of data that we need. In turn, we will make model training more efficient and sustainable. Even with existing data, intentional filtering and selection are crucial to ensure that a larger fraction of training mixture is of high quality.

As we continue to learn how to define the shape of data, and how these dimensions impact model performance, an evolution of this approach could play a role in active learning \citep{mindermann2022prioritized, evans2024data}, where models prioritize the right type of data during training via human-in-the-loop and model-in-the-loop, potentially accelerating progress even further. Moreover, the relation between topological dimensions of data and model performance is likely to provide us crucial pointers on where current learning paradigms fail, and hence inform the next generations of learning paradigms as well as relative value of various datasets.

By adopting a more intentional approach, we can build a more focused and efficient AI-powered future, using resources efficiently and paving the way for tackling complex AI challenges that require more than just data and scale.

\subsection*{Acknowledgments}
We are extremely grateful to Jennifer Prendki, Andrew Dai, Anoop Sinha, Blaise Aguera y Arcas, Isabelle Guyon, Rahul Sukthankar, Ethan Dyer, Rosanne Liu, Kenric Allado-McDowell, and Brendan Conway-Smith for their input and discussions.
\bibliography{references}

\end{document}